\pdfoutput=1
\documentclass{article}

\usepackage{microtype}
\usepackage{graphicx}
\usepackage{subfigure}
\usepackage{booktabs} % for professional tables
\usepackage{bm}
\usepackage{appendix}

\usepackage{hyperref}

\usepackage[accepted]{icml2018}

\icmltitlerunning{Benchmarking Automatic Machine Learning Frameworks}

\begin{document}

\twocolumn[
\icmltitle{Benchmarking Automatic Machine Learning Frameworks}



\icmlsetsymbol{equal}{*}

\begin{icmlauthorlist}
\icmlauthor{Adithya Balaji}{equal,gp}
\icmlauthor{Alexander Allen}{equal,gp}
\end{icmlauthorlist}

\icmlaffiliation{gp}{Georgian Partners, Toronto, Ontario, Canada}

\icmlcorrespondingauthor{Adithya Balaji}{abalaji2@ncsu.edu}
\icmlcorrespondingauthor{Alexander Allen}{arallen4@ncsu.edu}

\icmlkeywords{AutoML, automatic machine learning, TPOT, sklearn, auto-sklearn, H2O, auto\_ml}

\vskip 0.3in
]


\printAffiliationsAndNotice{\icmlEqualContribution}

\begin{abstract}
AutoML serves as the bridge between varying levels of expertise when designing machine learning systems and expedites the data science process. A wide range of techniques is taken to address this, however there does not exist an objective comparison of these techniques. We present a benchmark of current open source AutoML solutions using open source datasets. We test auto-sklearn, TPOT, auto\_ml, and H2O’s AutoML solution against a compiled set of regression and classification datasets sourced from OpenML and find that auto-sklearn performs the best across classification datasets and TPOT performs the best across regression datasets.
\end{abstract}

\section{Introduction}
The progression of machine learning from niche R\&D applications to enterprise applications creates a need for techniques that are accessible to companies that do not have the resources to hire an experienced data science team.

In response to the demand for accessible, automatic machine learning (AutoML) platforms, open source frameworks have been created to extract value from data as quickly and with as little effort as possible. These platforms automate most of the tasks associated with constructing and implementing a machine learning pipeline that would normally be engineered by specialized teams. AutoML platforms provide value to businesses who already have in-house data science teams and allow them to focus on more complex processes such as model construction without spending time on time-consuming processes such as feature engineering and hyperparameter optimization.

There are multiple areas of focus for automatic machine learning. There are a diverse set of AutoML frameworks claiming to produce the most valuable results with the least amount of effort. These frameworks apply relatively standardized techniques to the data developed over the years and collected in open source machine learning libraries such as scikit-learn. However, the methods that are used to automate the application and assessment of these techniques widely differ. These methods cannot be assessed on the rigor of their theory alone or by the individual performance of the constituent algorithms. Thus, they must be experimentally assessed as a whole across a variety of data. We perform a quantitative assessment on the most mature open source solutions available for AutoML.

\section{Selected Frameworks}
\subsection{Auto\_ml}
\href{https://github.com/ClimbsRocks/auto_ml}{Auto\_ml} is a framework designed to be used in production systems to allow companies to quickly pass extracted value from their data on to their customers. Auto\_ml automates many parts of a machine learning pipeline. First, it automates the feature engineering process through tf-idf processing (natural language), date processing, categorical encoding and numeric feature scaling. Its date preprocessing includes converting timestamps into binary features like weekend or weekday and splitting up components such as day, month and year. Auto\_ml also performs feature reduction when more than 100,000 columns exist using reduction methods such as PCA. This library requires the type of each feature as input in order to preprocess correctly. In addition, auto\_ml automates the model construction, tuning, selection, and ensembling process.

Auto\_ml utilizes highly optimized libraries such as Scikit-Learn, XGBoost, TensorFlow, Keras, and LightGBM for its algorithm implementations. It also contains pre-built model infrastructures for each classification and regression method which have a $<$ 1 millisecond prediction time. It optimizes models using an evolutionary grid search algorithm from sklearn-deap., 

Despite its features, it has poor extensibility. It also tends to performs poorly with multi-class classification problems. It also does not support a time limiting feature and thus each algorithm must be run to completion in an unbounded amount of time. This weighs against the usage of this framework in time constrained scenarios such as frequently retrained production systems. Also note that version 2.7.0 was used when testing this framework.

\subsection{Auto-sklearn}

\begin{figure}[ht]
\vskip 0.2in
\begin{center}
\centerline{\includegraphics[width=\columnwidth]{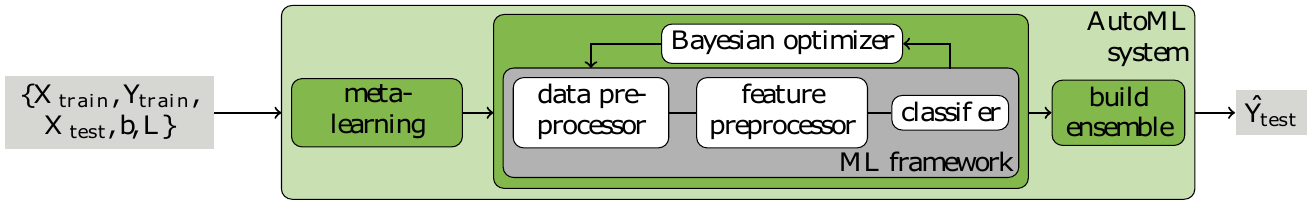}}
\caption{Auto-sklearn’s process for pipeline optimization}
\label{autosklearn}
\end{center}
\vskip -0.2in
\end{figure}

\href{https://github.com/automl/auto-sklearn}{Auto-sklearn} wraps the sklearn framework to automatically create a machine learning pipeline. It includes feature engineering methods such as one-hot encoding, numeric feature standardization, PCA and more. The models use sklearn estimators for classification and regression problems. Auto-sklearn creates a pipeline and optimizes it using bayesian search. The default hyperparameter values are warm started using 140 pre-trained datasets from OpenML using the meta-learning process outlined in figure \ref{autosklearn} \cite{NIPS2015_5872}. It computes 38 statistics for a dataset and initializes the hyperparameters to the optimized parameters of a dataset with statistics closest to the train set (determined by L1 distance).

Auto-sklearn tries all the relevant data manipulators and estimators on a dataset but can be manually restricted. It also has multi-threading support. One of the greatest advantages of this platform is its easy integration into the existing sklearn ecosystem of tools which provides an avenue for extension. Auto-sklearn uses the optimization framework SMAC3 which implements bayesian search along with a racing mechanism to quickly assess model performance.

This package lacks the ability to process natural language inputs and the ability to automatically discern between categorical and numerical inputs. The package also does not handle string inputs and requires manual integer encoding to accept categorical strings. Note that version 0.4.0 was used to test this framework.

\subsection{TPOT}

\begin{figure}[ht]
\vskip 0.2in
\begin{center}
\centerline{\includegraphics[width=\columnwidth]{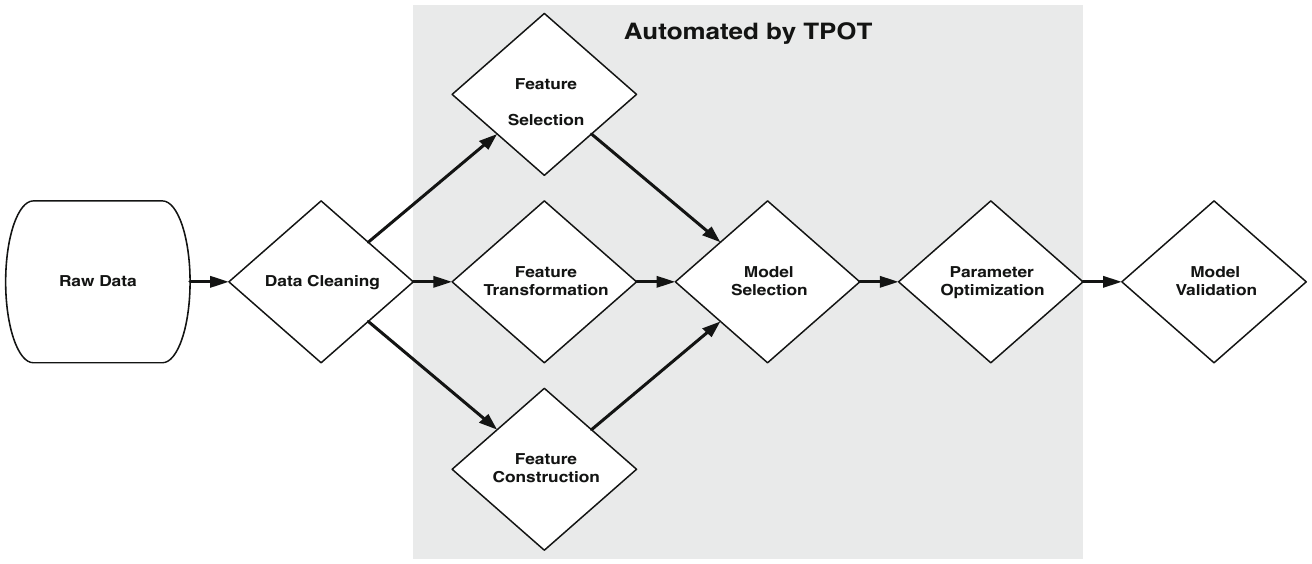}}
\caption{Pieces of the machine learning process automated by TPOT}
\label{TPOT}
\end{center}
\vskip -0.2in
\end{figure}

\href{https://github.com/EpistasisLab/tpot}{TPOT} or Tree-Based Pipeline Optimization Tool, is a genetic programming-based optimizer that generates machine learning pipelines. It extends the scikit learn framework with its own base regressor and classifier methods. It automates portions of the machine learning process detailed in figure \ref{TPOT} \cite{Olson2016EvoBio}.

Like auto-sklearn, TPOT sources its data manipulators and estimators from sklearn and its search space can be limited through a configuration file. Time restrictions are applied to TPOT by changing the maximum execution time or the population size. The optimization process also supports pausing and resuming. The most important feature of this framework is the ability to export a model to code to be further modified by hand.

TPOT cannot automatically process natural language inputs and also is not able to processes categorical strings which must be integer encoded before passing in data. Also note that version 0.9 was used when testing this framework.

\subsection{H2O}

\href{http://docs.h2o.ai/h2o/latest-stable/h2o-docs/automl.html}{H2O} is a machine learning framework similar to scikit-learn containing a collection of machine learning algorithms that execute on a server cluster accessible by a variety of interfaces and programming languages. h2o includes an automatic machine learning module that uses its own algorithms to build a pipeline. Configuration is limited to algorithm choice, stopping time, and degree of k-fold validation. It performs an exhaustive search over its feature engineering methods and model hyperparameters to optimize its pipelines. 

It currently supports imputation, one-hot encoding, and standardization for feature engineering and supports generalized linear models, basic deep learning models, gradient boosting machines, and dense random forests for its machine learning models. It supports two methods of hyperparameter optimization; cartesian grid search and random grid search. The end result is an ensemble model that can be saved and reloaded into the h2o framework to be used in production systems.

h2o is developed in Java and includes Python, Javascript, Tableau, R and Flow (web UI) bindings. The core code runs on a local or remote server to which external code connects and uploads jobs to be run. Production models are exported as native java entities that can be loaded into any h2o cluster.

The primary drawback of h2o is its massive resource usage. During testing, this framework suffered multiple failures during long-running processes due to inadequate garbage collection. Note that version 3.20.0.2 was used to test this framework.

\section{Benchmarking Methodology}
\subsection{Overview}
To accurately compare the selected AutoML frameworks, we design a \href{https://github.com/georgianpartners/automl_benchmark}{testing} rig to assess each framework’s effectiveness. We write snippets of code for each of the selected frameworks (TPOT, auto-sklearn, h2o, and auto\_ml) using their respective pipelines. Depending on the type of modeling problem, regression or classification, we use different metrics: MSE and F1 score (weighted), respectively.

The selection of the benchmarking datasets proves to be a challenging problem. Many open datasets require extensive preprocessing before use and do not come in the same shape or form. The OpenML database is chosen to solve this problem. \cite{OpenML} OpenML hosts datasets on their website while exposing an API to access the datasets. We use a custom set of 57 classification datasets and a custom set of 30 regression datasets to benchmark each framework., 

In order to achieve consistent results, we generate a set of 10 random seeds to fix the random number generator. This results in a compute space of 3,480 test items (10 seeds * 4 frameworks * 87 datasets). We set a soft compute time limit of 3 hours per framework and a hard limit of 3.5 hours based on a survey of each framework’s runtime. The combination of compute time and the search space results in an estimated 10,440 compute hours. This is infeasible to compute locally, thus we implement a distributed solution to execute the benchmarking suite.

\subsection{Distributed Computing Setup}
\subsubsection{AWS Batch}
We initially choose the Amazon Web Services (AWS) Batch framework to handle the parallelization and load balancing for this benchmark. This framework accepts a Docker container from an Amazon Elastic Container Repository (ECR). The container is then repeatedly executed on an Elastic Container Service (ECS) managed cluster of Elastic Compute (EC2) instances.

We configure the ECS compute cluster to create C4 compute-optimized EC2 instances. These instances run on Intel Xeon E5-2666 v3 processors operating at a 2.6 GHz base clock with a max clock of 3.3 Ghz. Amazon’s Hardware Virtual Machine (HVM) virtualization method is used to host these instances while Docker is used to host the individual containers. These instances include 1.88 Gb of memory per vCPU, 2 to 16 vCPUs and 250 Mbps storage bandwidth per vCPU. ECS automatically selects a combination of instances to achieve the necessary number of vCPUs to fully parallelize the process. Instances with more than 16 vCPU instances are available but consistency issues with Docker and Elastic Block Storage (EBS) arise when attempting to execute more than 16 containers on a single EC2 instance. Thus, we restrict the instance types available to ECS to the c4.4xlarge tier and below. We allocate 2 vCPUs and 4 GB of memory to each container and ECS handles the rest of the provisioning process.

We create a Docker image based on the Amazon Linux image because it is lightweight and compatible with AWS services. This image includes a script which bootstraps the container with all requirements at runtime. We also create a job file to serve as an index of all test items and upload it to S3. We then push the local repository to a remote branch and use the boto3 python framework to dispatch the AWS Batch array job to the compute cluster.

Upon container execution, the script clones down the repository, and provisions the python environment. The job file is also downloaded from S3 and parsed. A unique index passed into the container by AWS Batch determines which job to execute. Using an OpenML dataset number from the job, the required dataset is downloaded to the container using the OpenML API.

Finally, the benchmarking core code is called to execute the framework with the specified dataset, framework, and seed from the job file. If the framework overruns its time without generating a model we record this failure, kill the job, and move to the next test. We take a best-effort approach in ensuring all tests complete and all tests have at least 3 chances to succeed. The results are then uploaded to S3. These files are downloaded and concatenated locally to create the final output file which is used to generate the results.

\subsubsection{Bare Metal}
We find that AWS Batch managed compute environments and docker-based resource management can occasionally result in unpredictable behavior and performance on larger datasets on highly parallelized frameworks. The majority of h2o runs fail and many TPOT runs also fail due to memory limitations. Specifically, the docker manager sends a kill signal to the benchmarking process if the amount of memory used by the process exceeds the amount allocated by Batch. This hard limit is not differentiable from the memory used to allocate compute resources in Amazon ECS and thus cannot be changed without greatly increasing instance size per run. For AutoML frameworks that may spike in memory this is a major issue.

Additionally, it is a known issue that java does not respect docker container CPU and memory limits by default and thus the heap has to be manually allocated. However, as mentioned above, if the garbage collector space exceeds the max memory, the JVM is killed. This bug is the source of many of the failures of h2o. 

In response to these limitations of AWS Batch, we develop a custom distributed computing solution running on AWS. We spawn EC2 spot instances with boto3, provision them over SSH, then dispatch a task list to them. Instances are cleaned up by the dispatcher once the time limit for the processes are reached. This also allows us to allocate swap space which is necessary for h2o to execute on machines with small amounts of RAM. The amount of RAM makes no significant difference in the ability of these frameworks to execute as long as the limit is not reached. h2o more consistently succeeds using this method and is no longer killed for excessive RAM usage.

Although this mechanism is not containerized whereas AWS Batch is; the same memory and CPU constraints are applied to the system, which is executed on the same hypervisor and hardware platform and thus results in a identical operating environment. The virtualization layer of docker is proven to be negligible in regard to RAM access time, RAM space, and CPU capacity. \cite{7095802}

\subsection{Issues Encountered}
We encounter a multitude of issues when attempting to execute these automatic machine learning frameworks at scale that we do not typically see when executing them on single datasets. These issues arise when there are inconsistencies between the datasets we are using, inconsistencies between the random processes of the different pipelines, and inconsistencies between instances of the compute environments.

Some issues are in the datasets themselves. In some of the OpenML datasets, the “target” feature being predicted is null, a condition which none of the automatic machine learning frameworks are prepared to handle. Other edge cases include column name hash collisions and extremely large datasets such as the full MNIST set (70k data points) which none of the frameworks can complete within the required time and with the given resources.

Other issues exist in the random processes that occur in the machine learning pipeline. One common failure is in large multi-class classification tasks in which one of the classes lies entirely on one side of the train test split. Another case of this is when an entire category of a categorical variable lies on one side of the train test split. We also observe random failures with some of the sklearn estimators on specific seeds.

Finally, we resolve framework errors in TPOT and auto\_ml that prevent them from executing on certain datasets. In TPOT, the prediction method does not impute null values in the input, we resolve this by applying the default imputation method to any null values. Auto\_ml also has a number of bugs and dependency issues in regard to multi-class problems, and multi-model fitting. We also implement weighted F1 score metrics and optimization in auto\_ml.

\subsection{Individual Framework Configuration}

\begin{figure*}[!h]
\vskip 0.2in
\begin{center}
\centerline{\includegraphics[width=0.6\textwidth]{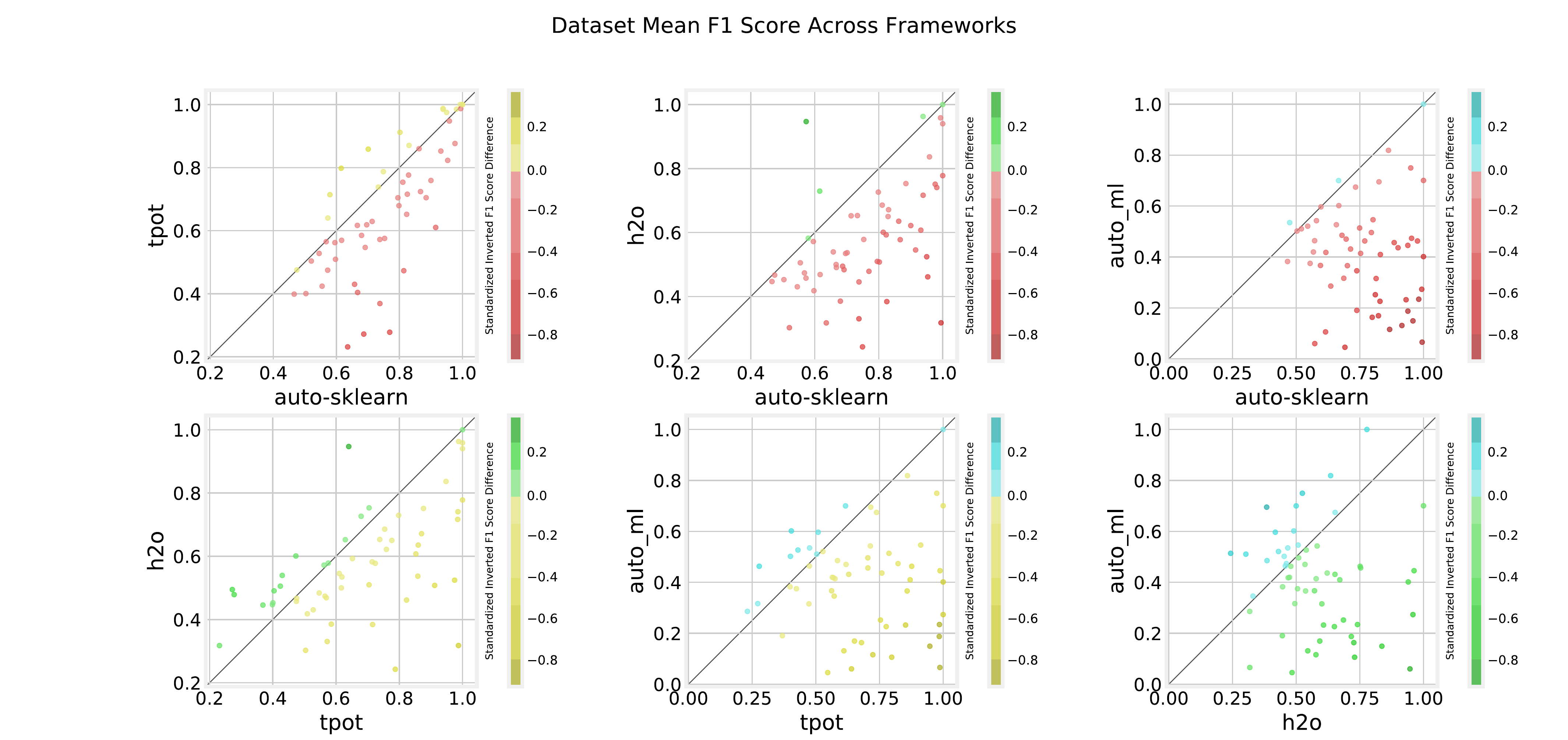}}
\caption{Framework head to head mean performance across classification datasets. Each chart is a one v one comparison of the performance of one framework with another. The axes represent the regularized F1 score of the frameworks.}
\label{pairwisef1}
\end{center}
\vskip -0.2in
\end{figure*}

\begin{figure*}[!h]
\vskip 0.2in
\begin{center}
\centerline{\includegraphics[width=0.6\textwidth]{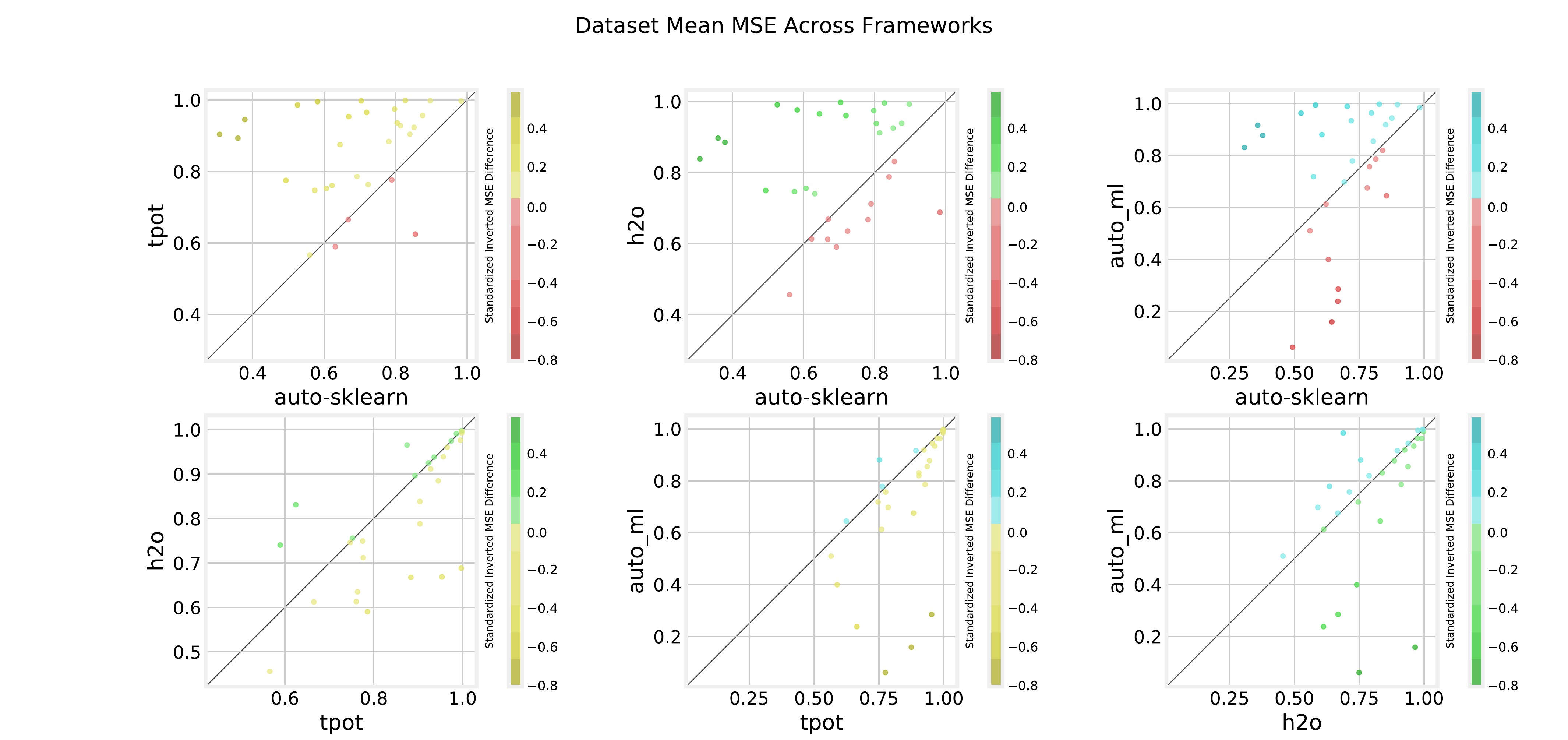}}
\caption{Framework head to head mean performance across regression datasets. Each chart is a one v one comparison of the performance of one framework with another. The axes represent the regularlized, scaled, and inversed RMSE (higher is better). Each data point is an average of 10 samples representing the framework run in the same configuration with 10 different, constant random seed values. Thus, each point is a representative sample of that frameworks performance on a dataset. Each data set processed is represented by a point.}
\label{pairwisemse}
\end{center}
\vskip -0.2in
\end{figure*}

We configure each framework as consistently as possible in order to ensure maximum fairness. Following is a brief summary of the configuration and preprocessing required to execute each framework.

For auto-sklearn, we set the total time left for the task to the total available runtime (3 hours) and we also set the per run time limit to an eighth of that value. The resampling strategy is set to five fold cross validation, and the optimization metric is either set to weighted F1 score or mean squared error depending on the problem type. In addition, we are required to provide whether each feature is a categorical column using OpenML metadata when fitting the estimator.

For TPOT we set the number of generations to 100 and the size of the population to 100. For classification problems, the internal LinearSVC estimator is disabled as it does not contain the predict\_proba function which is required for our scoring. We also set the max time to 3 hours, the optimization metric to weighted F1 or mean squared error, and the number of jobs to 2 (the number of vCPUs of the compute resource).

For h2o we set the number of threads to 2, the maximum runtime to 3.5 hours (this is decreased to 1.5 hours in increments of 0.5 hours to achieve maximum completion within the time limit), the minimum memory allocated to the JVM to seven gigabytes and the maximum memory allocated to the JVM to 100 gigabytes (enough to prevent capping out, this is provided through a swap partition allocated at VM provisioning). We also manually define the categorical columns as “factors” using OpenML metadata and set the optimization metric to mean squared error for regression and log-loss for classification (weighted F1 score is not implemented).

Finally, for auto\_ml we provide the feature type from the OpenML metadata and set the optimization metric to weighted F1 score for classification and mean squared error for regression. We also limit the classification estimators to AdaBoostClassifier, ExtraTreesClassifier, RandomForestClassifier and XGBClassifier, and limit the regression estimators to BayesianRidge, ElasticNet, Lasso, LassoLars, LinearRegression, Perceptron, LogisticRegression, AdaBoostRegressor, ExtraTreesRegressor, PassiveAggressiveRegressor, RandomForestRegressor, SGDRegressor and XGBRegressor. We are unable to set a time limit for auto\_ml and thus to remain within the time limits of the tests we disable GridSearchCV hyperparameter optimization. 

\section{Results}
\subsection{Dataset Analysis}

\begin{figure}[ht]
\vskip 0.2in
\begin{center}
\centerline{\includegraphics[width=1.15\columnwidth]{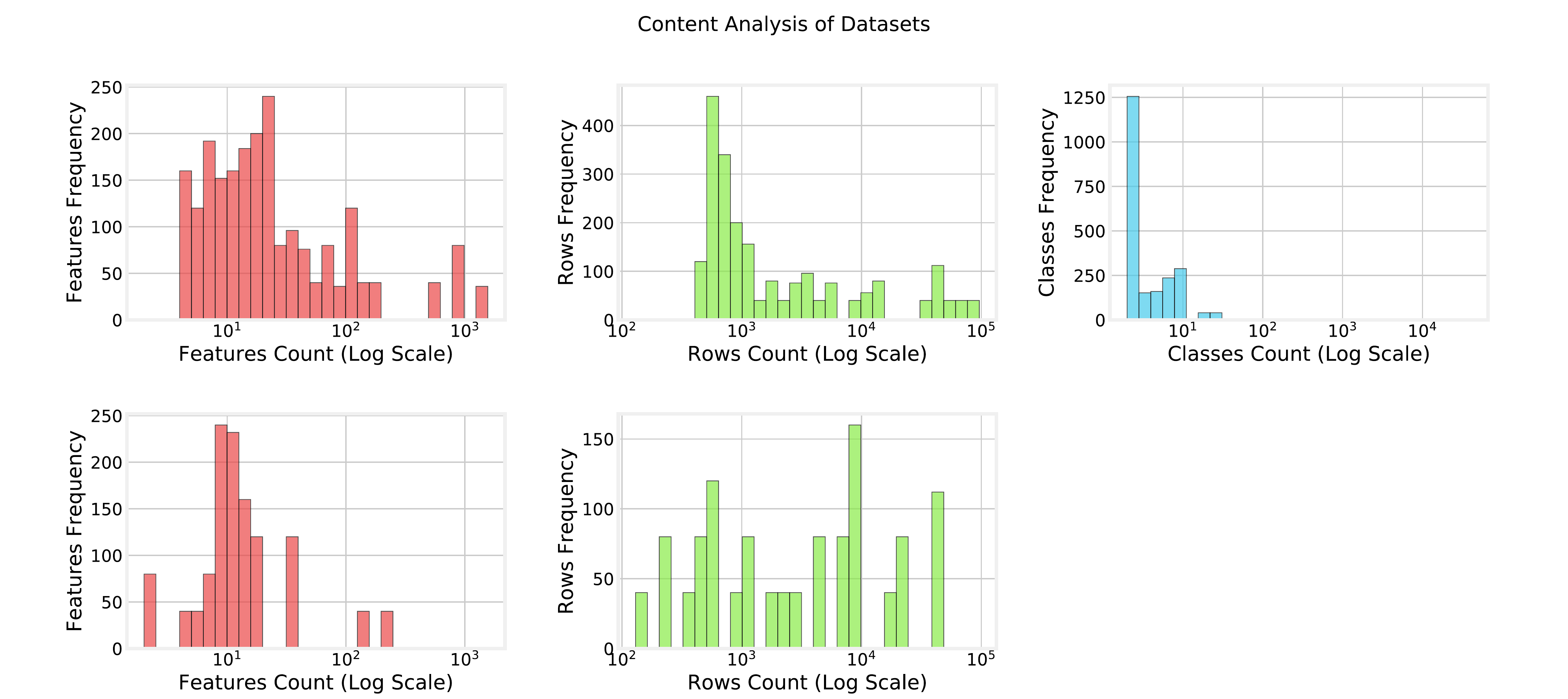}}
\caption{Raw dataset characteristics split between classification and regression problems. This figure shows the distribution of feature count, row count, and class count respectively left to right with the top figures being classification problems and the bottom being regression.}
\label{datadist}
\end{center}
\vskip -0.2in
\end{figure}

Our datasets are composed of 57 classification problems and 30 regression problems. We utilize a collection of diverse datasets sourced from OpenML to best assess all possible primary strength points of the chosen frameworks. It is important to note that each of the datasets are selected such that they were not internally used in any way by any of the AutoML methods. For example, auto-sklearn’s meta-learning is setup using datasets sourced from OpenML as well. The full set of chosen datasets is listed in the appendix tables 3 and 4.

Of the datasets chosen, we note that the data is not completely uniform and factors such as dataset size, feature size, and number of classification categories differ between datasets. We show here the biases present within our data and that those biases likely have no significant impact on the conclusions drawn from this point forward.

Figure \ref{datadist} demonstrates the general shape of the datasets, split between classification and regression tasks. We observe that classification primarily exists as a binary classification, regression row count is relatively uniform while classification row count is slightly skewed towards datasets around 1000 rows, and that feature count for both regression and classification center around 10 with classification skewing slightly towards 100 as well. Hence, we believe that this data group is a representative sample of general data science problems that many data scientists would encounter.

\subsection{Handling Missing Points}

\begin{table}[t]
\caption{Failure Count by framework}
\label{failures}
\vskip 0.15in
\begin{center}
\begin{small}
\begin{sc}
\begin{tabular}{lr}
\toprule
Framework    & Failure Count \\
\midrule
h2o          & 23            \\
tpot         & 6             \\
auto-sklearn & 0             \\
auto\_ml     & 0             \\
\bottomrule
\end{tabular}
\end{sc}
\end{small}
\end{center}
\vskip -0.1in
\end{table}

The count of failures for each framework is listed in table \ref{failures}. A total of 29 run combinations (dataset and seed) are dropped. These run combinations are dropped across all frameworks in order to maintain the comparability of frameworks. This process results in a total of 132 data points ($29*4$) that need to be dropped. This amounts to a drop percentage of ~3\% overall ($116 / 3480$ runs).

\subsection{Pairwise Framework Comparison}

Each framework is evaluated using the aforementioned split of regression and classification datasets. Performance is evaluated by aggregating the weighted F1 score and MSE scores across datasets by framework. It is important to note that each metric is standardized on a per dataset basis across frameworks and scaled from 0 to 1. In the case of MSE, these values were also inverted such that higher values represent better results so that the graphs would remain consistent between classification and regression visualizations. The mean across the 10 evaluated seeds is taken to represent a framework’s performance on a specific dataset.

In order to visualize the granular framework performance, the metrics are plotted in a pairwise manner as seen in figures \ref{pairwisef1} and \ref{pairwisemse}. Coloring is used to indicate the strength of a framework’s comparative performance. The stronger shaded points indicate greater the performance differences.

\subsection{Dataset Dependant Performance Analysis}

\begin{figure}[h!]
\vskip 0.2in
\begin{center}
\centerline{\includegraphics[width=1.15\columnwidth]{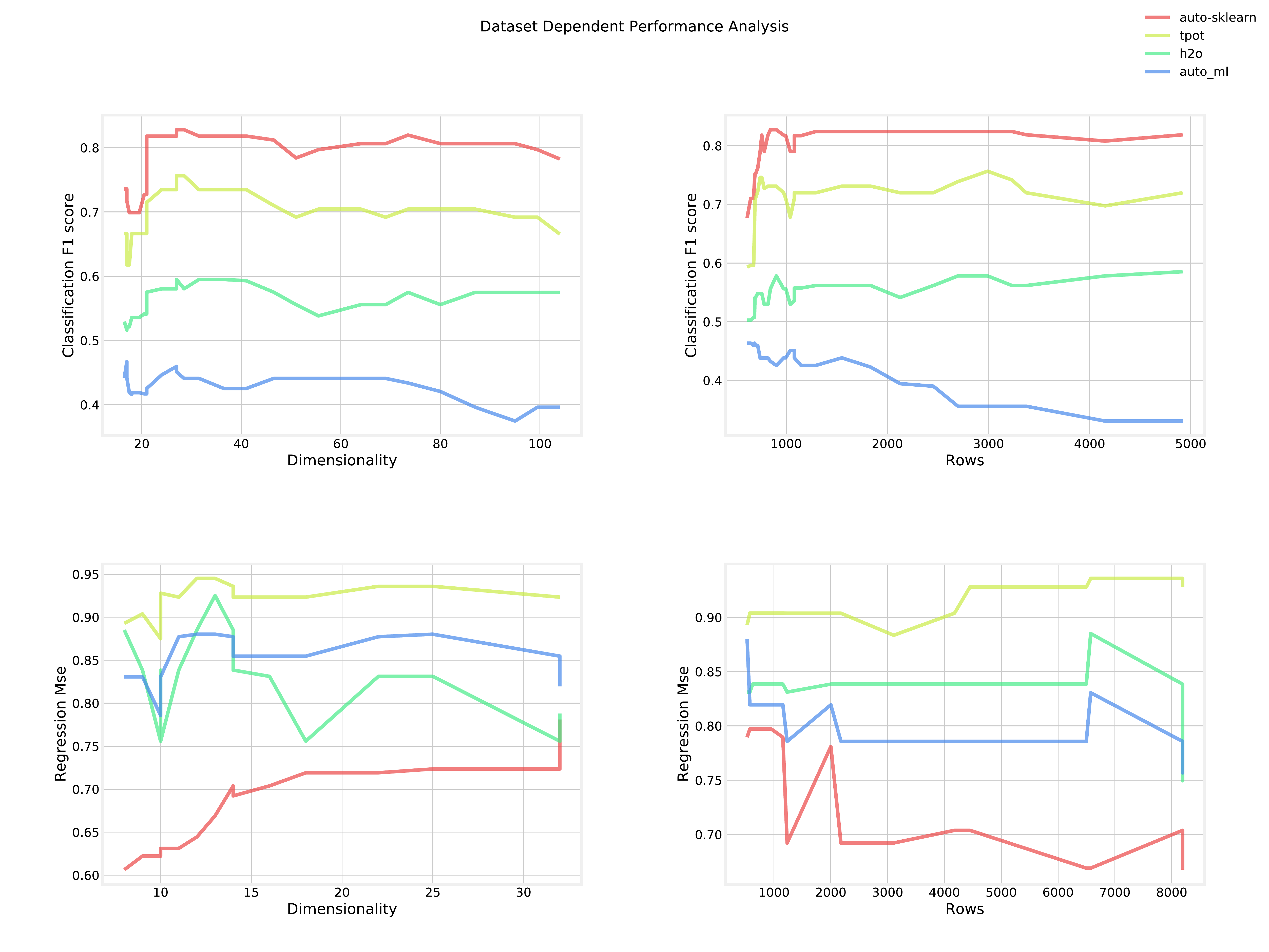}}
\caption{Mean performance per dataset compared against common dataset features. This figure shows a rolling average trend of framework performance versus dimensionality (left) and sample size (right) for each framework across classification (upper) and regression (lower) datasets.}
\label{depperf}
\end{center}
\vskip -0.2in
\end{figure}

Due to the natural clustering of the factors of our datasets, it is important to demonstrate that no innate bias exists towards frameworks that might have higher performance near those centers. We demonstrate this through figure \ref{depperf}, which compares the dimensionality and sample size to the corresponding performance metrics using a rolling average. The rolling average across the x-axis displays a more clear trend line and filters noise. Note that the same transformations applied in the pairwise comparison step are applied to MSE scores and weighted F1 scores in this case as well. One can see that in most cases framework performance is consistent across ranges of dimensionality and sample sizes. One notable outlier is auto\_ml’s regression tests which shows significant performance decrease on datasets with large rows compared to the rest of the frameworks. We attribute this degradation to the lack of hyperparameter optimization.

\subsection{Overall Comparison}

\begin{figure}[h!]
\vskip 0.2in
\begin{center}
\centerline{\includegraphics[width=\columnwidth]{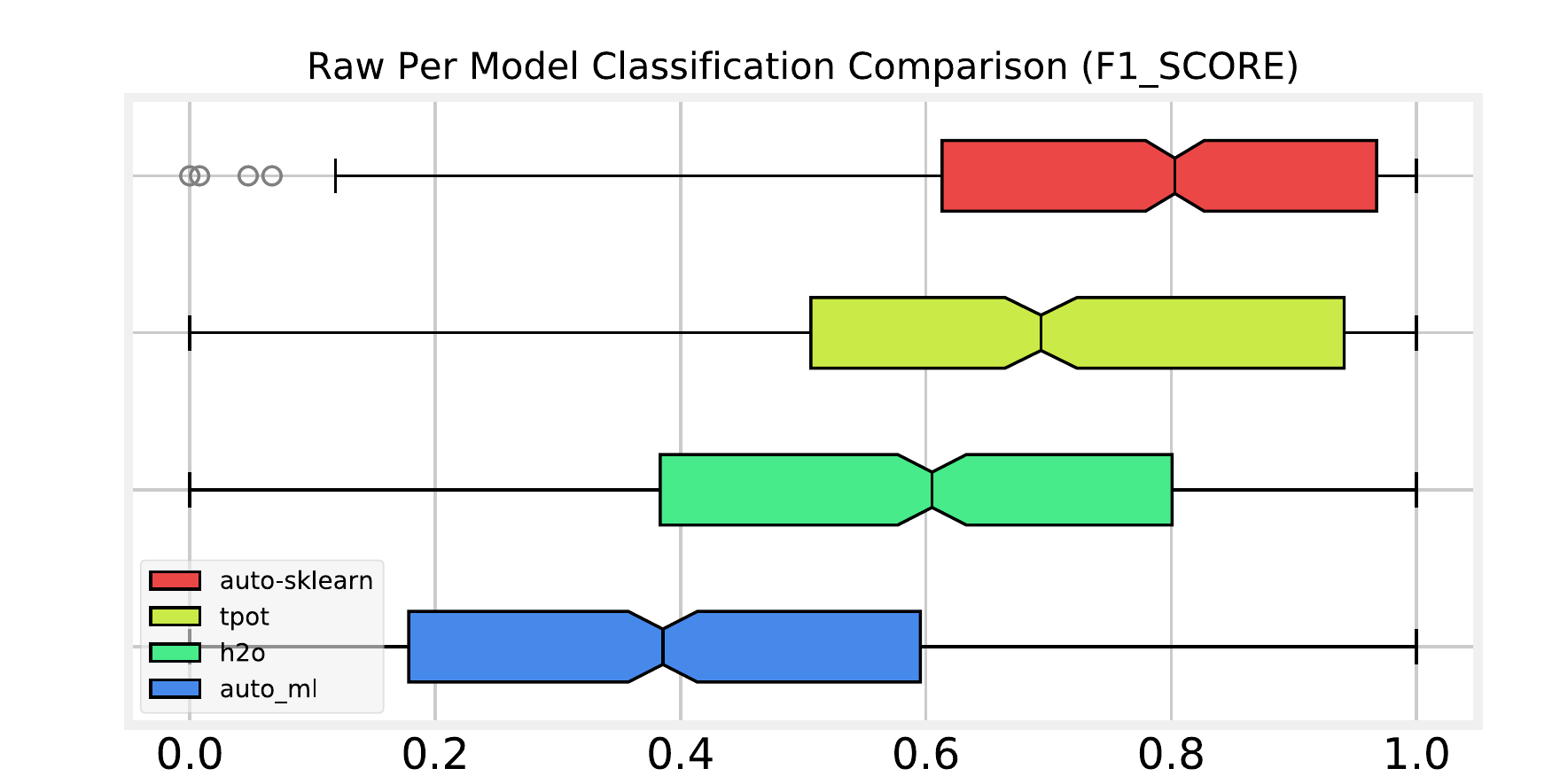}}
\caption{Framework performance across all classification datasets}
\label{fullf1}
\end{center}
\vskip -0.2in
\end{figure}

\begin{figure}[h!]
\vskip 0.2in
\begin{center}
\centerline{\includegraphics[width=\columnwidth]{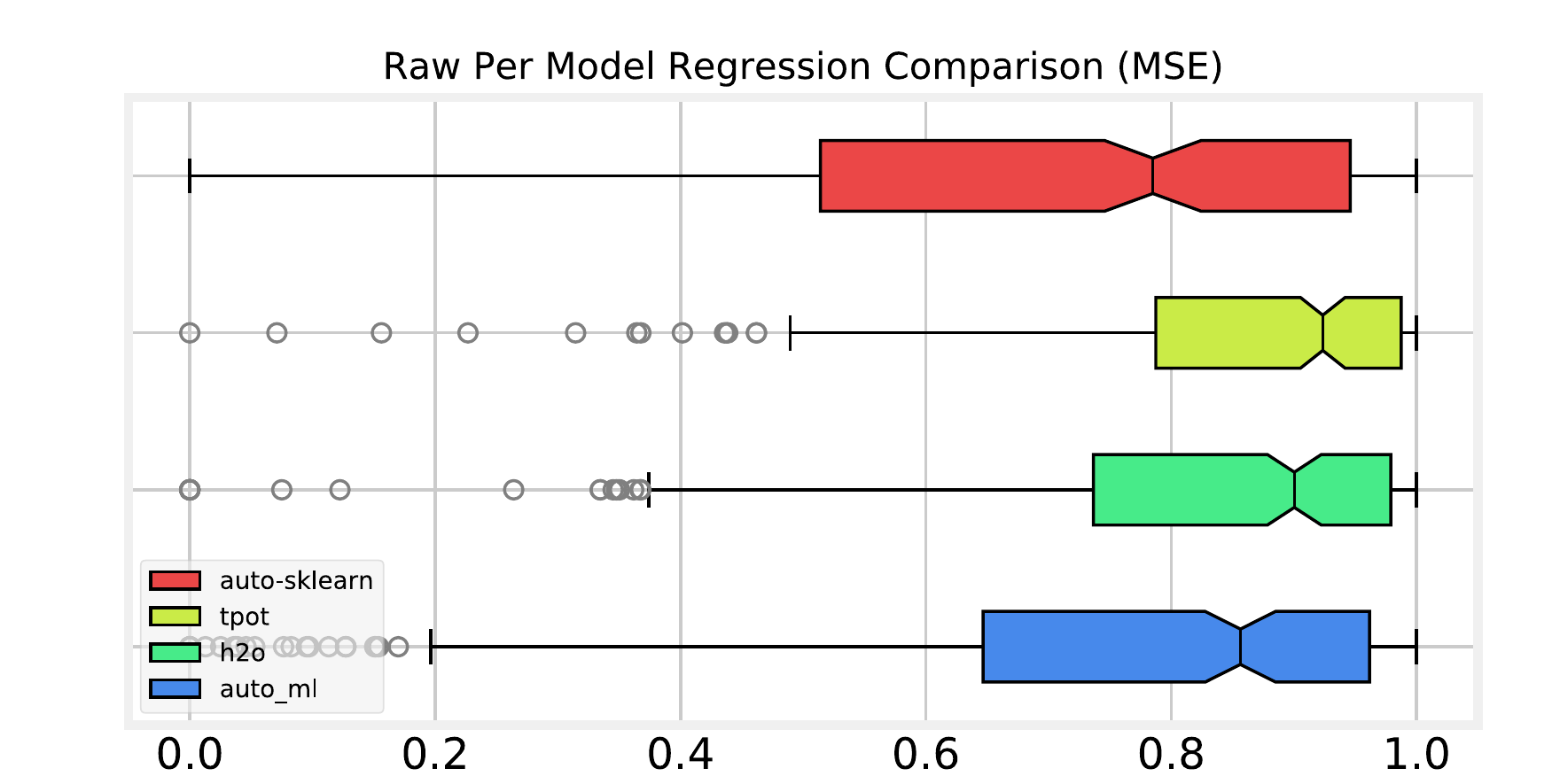}}
\caption{Framework performance across all regression datasets. These figures show the median performance (center line), quartiles (box and whisker), and 95\% median confidence range (notch in box), and potential outliers (grey circles) for each framework tested. }
\label{fullmse}
\end{center}
\vskip -0.2in
\end{figure}

Overall, auto-sklearn performs the best on the classification datasets and tpot performs the best on regression datasets. The same transformation of statistics from the pairwise comparison section are in use. We use boxplots to concisely demonstrate framework performance as seen in figures \ref{fullf1} and \ref{fullmse}. The notches in the plots represent the 95th confidence interval of the median. The unscaled means and variances for comparison are found in appendix tables 1 and 2.

\section{Conclusion and Recommendations}
We find that auto-sklearn performs the best on the classification datasets and TPOT performs the best on regression datasets. The quantitative results from this experiment have extremely high variances, as such, it is important to think carefully about the quality of the code base, activity, feature set, and goals of these individual frameworks. Potential future work includes more granular comparison of the specific feature engineering, model selection, and hyperparameter optimization techniques of these projects and to perhaps expand the scope to more AutoML libraries and frameworks as they are developed.

\bibliography{main.bbl}
% \bibliography{AutoMLBib}
\bibliographystyle{icml2018}

\clearpage
\begin{appendices}
\onecolumn
\section{Additional Tables}
\begin{table}[ht!]
\caption{Per framework scaled medians of results and their 95\% confidence intervals. Note MSE is inverted.}
\label{perframeworkmedians}
\vskip 0.15in
\begin{center}
\begin{small}
\begin{sc}
\begin{tabular}{lcr}
\toprule
Framework    & F1 Score & MSE \\
\midrule
auto-sklearn & $\bm{0.753 \pm 0.018}$         & $0.698 \pm 0.020$ \\
auto\_ml     & $0.420 \pm 0.018$         & $0.825 \pm 0.026$ \\
h2o          & $0.540 \pm 0.014$         & $0.835 \pm 0.024$ \\
tpot         & $0.679 \pm 0.022$         & $\bm{0.904 \pm 0.018}$ \\
\bottomrule
\end{tabular}
\end{sc}
\end{small}
\end{center}
%\vskip -0.1in
\end{table}

\begin{table}[h]
\caption{Per framework un-scaled means of results intervals. Note MSE is not inverted.}
\label{perframeworkmeans}
\vskip 0.15in
\begin{center}
\begin{small}
\begin{sc}
\begin{tabular}{lcr}
\toprule
Framework    & F1 Score & MSE \\
\midrule
auto-sklearn & 0.881         & $3.47e08$                \\
auto\_ml     & 0.849         & $1.20e08$                \\
h2o          & 0.862         & $1.04e08$                \\
tpot         & 0.875         & $1.06e08$ \\
\bottomrule
\end{tabular}
\end{sc}
\end{small}
\end{center}
% \vskip -0.1in
\end{table}

\begin{table}[H]
\caption{A list of each regression dataset used its OpenML id, name, and some general features.}
\label{regressionids}
\vskip 0.15in
\begin{center}
\begin{small}
\begin{sc}
\begin{tabular}{llrr}
\toprule
OpenML Dataset ID & Dataset Name     & Rows  & Features \\
\midrule
183               & abalone          & 4177  & 9        \\
189               & kin8nm           & 8192  & 9        \\
196               & autoMpg          & 398   & 8        \\
215               & 2dplanes         & 40768 & 11       \\
216               & elevators        & 16599 & 19       \\
223               & stock            & 950   & 10       \\
227               & cpu\_small       & 8192  & 13       \\
287               & wine\_quality    & 6497  & 12       \\
308               & puma32H          & 8192  & 33       \\
344               & mv               & 40768 & 11       \\
405               & mtp              & 4450  & 203      \\
495               & baseball-pitcher & 206   & 18       \\
497               & veteran          & 137   & 8        \\
503               & wind             & 6574  & 15       \\
505               & tecator          & 240   & 125      \\
507               & space\_ga        & 3107  & 7        \\
512               & balloon          & 2001  & 2        \\
528               & humandevel       & 130   & 2        \\
531               & boston           & 506   & 14       \\
537               & houses           & 20640 & 9        \\
541               & socmob           & 1156  & 6        \\
546               & sensory          & 576   & 12       \\
547               & no2              & 500   & 8        \\
549               & strikes          & 625   & 7        \\
550               & quake            & 2178  & 4        \\
558               & bank32nh         & 8192  & 33       \\
564               & fried            & 40768 & 11       \\
565               & water-treatment  & 527   & 37       \\
574               & house\_16H       & 22784 & 17       \\
41021             & Moneyball        & 1232  & 15     \\
\bottomrule
\end{tabular}
\end{sc}
\end{small}
\end{center}
% \vskip -0.1in
\end{table}

\begin{table}[h]
\caption{A list of each classification dataset used its OpenML id, name, and some general features.}
\label{classificationids}
\vskip 0.15in
\begin{center}
\begin{small}
\begin{sc}
\begin{tabular}{llrrr}
\toprule
ID & Dataset Name                                & Rows  & Features & Classes \\
\midrule

11                & balance-scale                               & 625   & 5        & 3       \\
15                & breast-w                                    & 699   & 10       & 2       \\
29                & credit-approval                             & 690   & 16       & 2       \\
37                & diabetes                                    & 768   & 9        & 2       \\
42                & soybean                                     & 683   & 36       & 19      \\
50                & tic-tac-toe                                 & 958   & 10       & 2       \\
54                & vehicle                                     & 846   & 19       & 4       \\
151               & electricity                                 & 45312 & 9        & 2       \\
188               & eucalyptus                                  & 736   & 20       & 5       \\
307               & vowel                                       & 990   & 13       & 11      \\
333               & monks-problems-1                            & 556   & 7        & 2       \\
334               & monks-problems-2                            & 601   & 7        & 2       \\
335               & monks-problems-3                            & 554   & 7        & 2       \\
375               & JapaneseVowels                              & 9961  & 15       & 9       \\
377               & synthetic\_control                          & 600   & 61       & 6       \\
451               & irish                                       & 500   & 6        & 2       \\
458               & analcatdata\_authorship                     & 841   & 71       & 4       \\
469               & analcatdata\_dmft                           & 797   & 5        & 6       \\
470               & profb                                       & 672   & 10       & 2       \\
1038              & gina\_agnostic                              & 3468  & 971      & 2       \\
1046              & mozilla4                                    & 15545 & 6        & 2       \\
1063              & kc2                                         & 522   & 22       & 2       \\
1459              & artificial-characters                       & 10218 & 8        & 10      \\
1461              & bank-marketing                              & 45211 & 17       & 2       \\
1462              & banknote-authentication                     & 1372  & 5        & 2       \\
1464              & blood-transfusion-service-center            & 748   & 5        & 2       \\
1467              & climate-model-simulation-crashes            & 540   & 21       & 2       \\
1468              & cnae-9                                      & 1080  & 857      & 9       \\
1476              & gas-drift                                   & 13910 & 129      & 6       \\
1479              & hill-valley                                 & 1212  & 101      & 2       \\
1480              & ilpd                                        & 583   & 11       & 2       \\
1485              & madelon                                     & 2600  & 501      & 2       \\
1486              & nomao                                       & 34465 & 119      & 2       \\
1510              & wdbc                                        & 569   & 31       & 2       \\
1515              & micro-mass                                  & 571   & 1301     & 20      \\
1590              & adult                                       & 48842 & 15       & 2       \\
4550              & MiceProtein                                 & 1080  & 81       & 8       \\
6332              & cylinder-bands                              & 540   & 38       & 2       \\
23380             & cjs                                         & 2796  & 34       & 6       \\
23381             & dresses-sales                               & 500   & 13       & 2       \\
23517             & numerai28.6                                 & 96320 & 22       & 2       \\
40496             & LED-display-domain-7digit                   & 500   & 8        & 10      \\
40499             & texture                                     & 5500  & 41       & 11      \\
40536             & SpeedDating                                 & 8378  & 121      & 2       \\
40668             & connect-4                                   & 67557 & 43       & 3       \\
40670             & dna                                         & 3186  & 181      & 3       \\
40701             & churn                                       & 5000  & 21       & 2       \\
40966             & MiceProtein                                 & 1080  & 78       & 8       \\
40971             & collins                                     & 1000  & 20       & 30      \\
40975             & car                                         & 1728  & 7        & 4       \\
40978             & Internet-Advertisements                     & 3279  & 1559     & 2       \\
40981             & Australian                                  & 690   & 15       & 2       \\
40982             & steel-plates-fault                          & 1941  & 28       & 7       \\
40983             & wilt                                        & 4839  & 6        & 2       \\
40984             & segment                                     & 2310  & 17       & 7       \\
40994             & climate-model-simulation-crashes            & 540   & 19       & 2       \\
41027             & jungle\_chess\_2pcs\_raw\_endgame\_complete & 44819 & 7        & 3     \\ 
\bottomrule
\end{tabular}
\end{sc}
\end{small}
\end{center}
\vskip -0.1in
\end{table}
\end{appendices}
\end{document}